\documentclass[10pt,a4paper]{IEEEtran}

\usepackage{amssymb,amsmath}
\usepackage[dvips]{graphicx} 
\usepackage{slashbox} 
\usepackage{tabularx}
\usepackage{cite}

\newcommand\T{\rule{0pt}{2.6ex}}
\newcommand\B{\rule[-1.2ex]{0pt}{0pt}}

\author{Wu Jiang, Fei Ding and Qiao-liang Xiang \\Department of Optoelectronic Engineering \\Nanjing University of Posts and Telecommunications \\ \texttt{\{albert.w.jiang, danix800, qiaoliangxiang\}@gmail.com}}
\title{An Affinity Propagation Based method for Vector Quantization Codebook Design}

\begin{document}
\maketitle

\begin{abstract}
 In this paper, we firstly modify a parameter in affinity propagation~(AP) to improve its convergence ability, and then, we apply it to vector quantization~(VQ) codebook design problem. In order to improve the quality of the resulted codebook, we combine the improved AP~(IAP) with the conventional LBG algorithm to generate an effective algorithm call IAP-LBG. According to the experimental results, the proposed method not only improves its convergence abilities but also is capable of providing higher-quality codebooks than conventional LBG method.\\
\end{abstract}

\section{Introduction}
Vector quantization~(VQ) is an effective method in lossy data compression, and is widely used in the field of speech coding, image coding, video compression, etc.\cite{ieeetr:Gersho}\cite{ieeetr:Gray}\cite{ieeetr:Nasrabadi}. According to the VQ processes, we know that, for image compression, the quality of the reconstructed image highly depends on the quality of the codebook. Hence the codebook design is a very important task for VQ. Essentially, the codebook design is a clustering problem. Given a set of training vectors $\mathbf{V} = \big\{ V_{n}: n = 1, 2, \ldots, N \big\}$, the goal is to search for a map $: \mathbf{V}: \big\{ V_{n}: n = 1, 2, \ldots, N \big\} \rightarrow \mathbf{C}: \big\{C_{m}: m = 1, 2, \ldots, M \big\}, N \gg M$ which maps each training vector~(or training point) $V_{n}$ to its cluster centroid $C_{m}$. This map shall minimize the cost function. Here, a simple squared Euclidean distance measure $E = \Sigma\lVert V_{n} - C_{m} \rVert^{2}$ is used as the cost function.\\


A generalized algorithm was proposed by Linde, Buzo, and Gray~(LBG)\cite{ieeetr:Linde}. It
is the most popular codebook design method. LBG iteratively applies two optimality conditions~(nearest neighbor condition and centroid condition) to generate a codebook. However, it suffers from local optimality and is sensitive to the initial solution. If the initial solution is poor, the resulted codebook's quality will probably be poor, and as a result it will be difficult to produce a high-quality image.\\

Recently, a powerful algorithm called Affinity Propagation (AP) for unsupervised
clustering was proposed by Frey and Dueck\cite{ieeetr:Frey} . In AP algorithm, each point in a set is viewd as a node in a network. AP is based on message passing along edges
of the network, following the idea of belief propagation \cite{ieeetr:Yedid}\cite{ieeetr:Yedidia}. AP takes input real-value similarities $s(n,m)$ which indicate how well the data point $m$ is suited to be the cluster centroid to data point $n$, and then, two kinds of real-value messages \emph{``responsibility''} $r(n,m)$ and \emph{``availability''} $a(n,m)$ are exchanged among data points until a high-qulity set of cluster centroids and corresponding clusters gradually emerges\cite{ieeetr:Frey}. Breifly, there are two significant advantages of AP: one is its high-quality clustering capabilty; the other is its computational efficiency, especially for large data sets\cite{ieeetr:Michele}. However, in AP, for \emph{self-similarity} is the same for each point, all data points are simultaneously considered as potential clustering centroids. Actually, this feature brings a drawback for AP, since it will be more difficult to converge.\\

In this paper, we propose an improved AP~(IAP) algorithm by modifying a parameter called \emph{network-support similarity} $ns(m,m)$ for each point which improves the algorithm's convergence abilities. In the original AP, all the \emph{self-similarities}\cite{ieeetr:Frey} are the same for all points,however, \emph{network-support similarity} $ns(m,m)$ for each point changes according to the network supports. As to a point in the network, \emph{network-support similarity} equals to the average squared Euclidean distance from this point to the other points in the whole network. We consider that data points with larger values of \emph{network-support similarities} are more probably to be chosen as clustering centroids. Also we offer a parameter called \emph{ratio of similarity} $rs$ to control the number of codewords needed. In addition, based on IAP algorithm, we propose an algorithm called IAP-LBG to effectively design the VQ codebook. Due to the strong clustering ability of IAP, the codebook's quality is further improved.\\

\section{LBG Algorithm}
Suppose that we want to design a codebook with $M$ codewords from a training set
$\mathbf{V}= \big\{ V_{n} : n = 1,2, \ldots, N \big\}$, where $ N \gg M $. The processes of the LBG algorithm are described as follows.

\begin{description}
 \item[Step 1] Randomly select $ M $ vectors from the training set $ \mathbf{V} $ as the initial codebook $ \mathbf{C} = \big\{ C_m : m = 1, 2, \ldots, M \big\}$;

 \item[Step 2] For $\forall V_{n} \in \mathbf{V}$, find the closest codeword in the current codebook $\mathbf{C}$ according to the squared Euclidean distance $E = \lVert V_{n} - C_{m} \rVert$ and then add the training vectors into the corresponding cluster of the closest codeword found.

 \item[Step 3] For $\forall C_{m} \in \mathbf{C}$, calculate the overall average distortion $D = \frac{1}{J \times M} \sum_{m=1}^{M} \sum_{j=1}^{J} E$ between the codeword and each training vector of the associated cluster. $ J $ is the total number of training vectors in the associated cluster.

 \item[Step 4] If the difference in overall average distortion between the last two successive iterations is smaller than some threshold or a certain number of iterations is reached, then terminate the iteration.

 \item[Step 5] For each codeword in the current codebook $\mathbf{C}$, evaluate the centroid of its associated cluster and take the centroid as a new codeword for the next iteration. Go back to Step 2.
\end{description}

The conventional LBG algorithm suffers from local optimality and is sensitive to the initial solution. So if the initial solution is poor, the generated codebook’s quality will probably be poor, and as a result it will be difficult to produce a high-quality image when decodeing.

\section{Affinity Propagation Algorithm}
AP takes as input similarities $ s(n,m) $ which indicate how well the data point $ m $ is suited to be the centroid for data point $ n $.  Here the similarity is set to be a negative squared Euclidean distance: 
\begin{equation}
 s(n,m) = -\lVert V_{n} - C_{m} \rVert ^{2} \label{fifth}. 
\end{equation}

$ s(m,m) $ indicates that data points with larger values are more likely to be chosen as clustering centroids. The number of the final examplars is influenced by the value of $ s(m,m) $. In the conventional AP, all data points are simultaneously considered as potential examplars so the authors set all $ s(m,m) $ to be the same value \cite{ieeetr:Frey}.\\

In the processing, two kinds of message are exchanged among data points, and each takes into account a different kind of competition. The \emph{``responsibility''} $ r(n,m) $, sent from data point $ n $ to candidate exemplar point $ m $, reflects the accumulated evidence for how well-suited point $ m $ is to serve as the exemplar for point $ n $, taking into account other potential exemplars for point $ n $. The \emph{``availability''} $ a(n,m) $ sent from candidate exemplar point $ m $ to point $ n $, reflects the accumulated evidence for how fitting it would be for point $ n $ to choose point $ m $ as its exemplar, taking into account the support from other points that point $ m $ should be an exemplar. To begin with, the availabilities are initialized to zero, and in the whole process, they followes the updating rule below.\\

\begin{subequations}\label{grp}
\begin{align}
r(n,m)&= s(n,m) - \max_{m^{'} \textrm{s.t.}\, m^{'} \neq m} \big\{ a(n,m^{'}) + s(n,m^{'}) \big\} \label{first}\\
r(m,m)&= s(m,m) - \max_{m^{'} \, \textrm{s.t.}\, m^{'} \neq m}\big\{ a(m,m^{'}) + s(m,m^{'}) \big\}\label{second} \\
a(n,m)&= \min \big\{0, r(m,m) + \sum_{n^{'} \textrm{s.t.}\, n^{'} \notin {n,m}} {\max \big\{0, r(n^{'},m)}\big\} \label{third}\\
a(m,m)&= \sum_{n^{'} \textrm{s.t.}\,n^{'} \neq m}{\max \big\{ 0,r(n^{'},m)\big\}}\label{forth}
\end{align}
\end{subequations}

Messages are updated on the basis of simple formula that search for minima of an appropriately chosen energy function. At any point in time, the magnitude of each message reflects the current affinity that that one data point has for choosing another data point as its exemplar. \\

For point $ n $, the value of $ m $ that maximizes $ a(n,m) + r(n,m) $ either identifies
point $ n $ as an exemplar if $ m = n $, or identifies the data point that is the
exemplar for point $ n $ \cite{ieeetr:Frey}. The message-passing procedure may be terminated after a fixed number of iterations, after changes in the messages fall below a thereshold, or after the local decisions stay constant for some number of iterations.\\

\section{Proposed Algorithm}
Since in the conventional AP, the authors consider that all data points can be equally suitable as exemplars, they set \emph{self-similarities} of each point to be the same. However, we propose a different view of $ s(m,m) $. We argue that the \emph{self-similarity} of each point should vary according to the similarities between this point and the others. A point may ``love'' to take itself as a exemplar more if it ``knows'' there are more other points choosing it to be a exemplar. We call this rule \emph{network-support similarities} which, in this paper, is denoted as $ ns(m,m) $:

\begin{equation}
  ns(m,m) = \frac{{\displaystyle \sum_{m^{'} \textrm{s.t.}\,m^{'} \neq m} \big\{ s(m^{'},m)\big\}}}{ N-1 } \label{sixth}
\end{equation}

We consider that the point whose $ns(m,m)$ is larger would be more appropriate to be an examplar. Because the cluster shape is regular in VQ codebook design, there is only one centroid for each cluster. As to a point, when more points support it to be a centroid, it should prefer to choosing itself as a centroid than other points. In order to get the very number of codewords, we set a tuning parameter called \emph{ratio of network-support similarities} $rs$. And we find that the codeword number decreases monotonously with $rs$.

\begin{equation}
 s(m,m) = rs \times ns(m,m) \label{seventh}
\end{equation}


To fine-tune the final solution, we use LBG algorithm after the process of the IAP. Since some codewords may be replaced, we must update the associated clusters to reduce distortion errors. Because IAP could generate good codebooks in the first step, the following LBG process would generate effective codebooks finally.\\



The process of the proposed algorithm is as follows.
\begin{description}
 \item[Step 1] Caculate \emph{similarities} $ s(n,m) $ for all data points according to Eq.\eqref{fifth}.
 \item[Step 2] Caculate \emph{network-support similarities} $ns(m,m)$ for all data points according to Eq.\eqref{sixth}, and initialize $rs=0.1$.
 \item[Step 3] Initialize all $a(n,m) = r(n,m) = 0$.
 \item[Step 4] $\forall n \in \big\{ 1:N \big\}$, update the $N$ \emph{responsibilities} $r(n,m)$ and then the $N$ \emph{availabilities} $a(m,n)$ parallelly according to Eqs.\eqref{grp}.
 \item[Step 5] Identify the exemplars $C_{m}$ by looking at the maximum value of $r(n,m) + a(n,m)$ for given $n$.
 \item[Step 6] Repeat Steps 4-5 until there is no change in exemplars for a large number of iterations.
 \item[Step 7] Modify $rs$ and repeat Step 3-6 to get the codebook of the right size.
 \item[Step 8] Use the codebook generated in Step 6 as an initial codebook then continue to use LBG method to generate the finial codebook.
\end{description}

Compared with the original AP, firstly, our improved AP~(IAP) could converge much faster~(see Figure 1). Secondly, our IAP-LBG algorithm could performance better on generating codebooks for VQ~(see Table 1 and Table 2). 

\section{Experimental Results}
Five $256 \times 256$ pixels monochrome images~(``peppers'', ``lena'', ``bridge'', ``camera'', and ``baboon'') with $256$ gray levels are used to evaluate the effectiveness of the proposed algorithm. Firstly, we train respective codebooks of size $ 256 $ with $16$ dimensions using corresponding images, then we take the codebook of image ``peppers'' as a universal codebook and use all images to test its quality. In addition, in the process of generating the codebook of ``peppers'', we also compare the convergence abilities of IAP with the conventional AP algorithm.\\

Performance comparisons are made among the conventional LBG algorithm\cite{ieeetr:Linde}, the conventional AP algorithm\cite{ieeetr:Frey}, the IAP algorithm and the IAP-LBG algorithm. For LBG, the maximum number of iterations is set to be $50$, and threshold is $10^{-8}$ and the final results are obtained after $ 20 $ runs. For AP, IAP and IAP-LBG, when updating the message, we also use a parameter called \emph{damping factor}\cite{ieeetr:Frey} to avoid numerical oscillations that arise in some circumstances. The value of damping factor is set to be $0.5$ in all of our experiments. As to AP, IAP and IAP-LBG, since $rs$ is modified by the characters of the training sets, $rs$ varies as image changes. We tune the parameter $rs$ to obtain $256$-size codebooks. They are given as follows, $ rs(peppers) = 0.086, rs(lena) = 0.14, rs(camera) = 0.149, rs(baboon) = 0.280, rs(bridge) = 0.203$.\\

Firstly, We compare the convergence abilities of IAP with the conventional AP algorithm. From Figure 1, we can clearly see that it only takes IAP $71$ iterations to converge, however, it takes AP more than $100$ iterations. Moreover, the value of the energy function is almost the same.\\

\begin{figure}[h]
 \centering
 \includegraphics[width=0.5\textwidth]{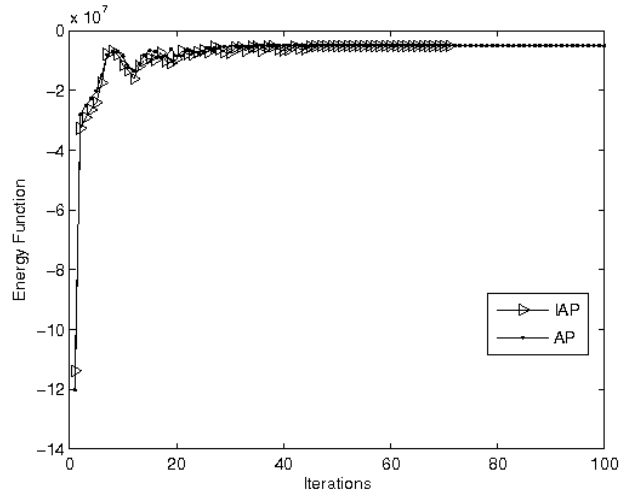}
 \caption{ \textnormal{Comparisons on convergence abilities of AP and IAP} }
 \label{fig:graph}
\end{figure}

Comparisons measured by PSNR~(dB) on genarating codebooks for the five different images are compared among the four methods. Results are shown in Table 1 and Table 2. The codebooks used in Table 1 are generated from the training sets accordingly, and the codebook used in Table 2 is generated from the training set of the ``peppers''. From Table 1, we can see that IAP-LBG method can improve the PSNR of the generated codebook by $ 0.62 $ dB compared with conventional AP, and $ 0.95 $ dB compared with conventional LBG averagely. From Table 2 we can see that IAP-LBG algorithm can improve the PSNR by $ 0.18 $ compared with conventional AP, and $0.28$ compared with conventional LBG averagely. In a word, the proposed algorithm in this paper is really effective.

 
\begin{table}[h]
\begin{center}
\begin{tabular}{|c|c|c|c|c|}
\hline
Algorithms\T \B &LBG &AP& IAP & IAP-LBG \\\hline\hline
peppers \T \B & 31.18 & 31.46& 31.38 & 32.04 \\\hline
lena\T \B & 29.67 & 30.02& 29.94 & 30.64 \\\hline
camera \T \B &25.95 & 27.68&27.69 & 28.49 \\\hline
baboon \T \B & 26.52 &26.11& 26.03 & 26.71 \\\hline
bridge \T \B &25.48 & 25.18& 25.07 & 25.67 \\\hline
average\T \B & 27.76 & 28.09& 28.02 & 28.71 \\\hline
\end{tabular}
\end{center}
\caption{\textnormal{PSNR~(dB) within training set}}
\end{table}

\begin{table}[h]
\begin{center}
\begin{tabular}{|c|c|c|c|c|}\hline
Algorithms\T \B &LBG &AP& IAP & IAP-LBG \\\hline\hline
peppers \T \B & 31.18 &31.46& 31.38 & 32.04 \\\hline
lena \T \B & 27.34 & 27.40& 27.39 &27.54\\\hline
camera \T \B & 22.55 &22.79& 22.57 & 22.75 \\\hline
baboon \T \B & 24.97 &24.89& 25.01 & 25.05 \\\hline
bridge \T \B & 23.65 &23.68& 23.66 & 23.74\\\hline
average \T \B & 25.94 &26.04 & 26.00 & 26.22\\\hline
\end{tabular}
\end{center}
\caption{\textnormal{PSNR~(dB) outside training set}}
\end{table}

\section{Conclusions}
In this paper we propose a method called IAP-LBG which improves the quality of VQ codebook. Firstly we improve the convergence abilities of the conventional AP algorithm by modifying a parameter called \emph{network-support similarities}, then take this codebook as initial codebook and use the conventional LBG method to generate a high-quality codebook. In the experiment, performance comparisons made among five different images potently proved its effectiveness in reconstructed images quality. \\


\section{Acknowledgements}
We acknowledge very useful discussions with Yu-Xuan~Wang, and we are grateful to ECHO lab for providing us with the facilities we need to finish our work.
\bibliographystyle{ieeetr}
\bibliography{reference}

\begin{thebibliography}{1}

\bibitem{ieeetr:Gersho}
A.~Gersho and R.~M. Gray, {\em Vector Quantization and Signal Compression}.
\newblock Kluwer Academic Publishers, 1992.

\bibitem{ieeetr:Gray}
R.~M. Gray, ``Vector quantization,'' {\em IEEE Acoustics, Speech, Signal
  Processing Magazine, Vol. 1, pp. 4-29}, April 1984.

\bibitem{ieeetr:Nasrabadi}
N.~M. Nasrabadi and R.~A. King, ``Image coding using vector quantization: A
  review,'' {\em IEEE Transactions on Communications, Vol. 36, pp. 957-971},
  1988.

\bibitem{ieeetr:Linde}
Y.~Linde, A.~Buzo, and R.~M. Gray, ``An algorithm for vector quantizer
  design,'' {\em IEEE Transactions on Communications, Vol. 28, pp. 84-95},
  January 1980.

\bibitem{ieeetr:Frey}
B.~J. Frey and D.~Dueck, ``Clustering by passing messages between data
  points,'' {\em Science Magazine, Vol. 315, pp. 972-976}, February 2007.

\bibitem{ieeetr:Yedid}
J.~S. Yedidia, W.~T. Freeman, and Y.~Weiss, ``Understanding belief propagation
  and its generalizations,'' {\em Mitsubishi Electric Research Laboratories},
  2002.

\bibitem{ieeetr:Yedidia}
J.~S. Yedidia, W.~T. Freeman, and Y.~Weiss, ``Belief propagation and
  generalizations,'' {\em IEEE Transactions on Information Theory, Vol. 51, pp.
  2282}, 2005.

\bibitem{ieeetr:Michele}
M.~Leone, Sumedha, and M.~Weight, ``Clustering by soft-constraint affinity
  propagation: Applications to gene-expression data,'' {\em arXiv:
  0705.2646vl}, 2007.

\end{thebibliography}
\end{document}